\documentclass[10pt,twocolumn,letterpaper]{article}

\usepackage[pagenumbers]{cvpr} 
\usepackage{tikz}
\usetikzlibrary{shapes.geometric, arrows.meta, positioning, shadows, calc}
\usepackage{stfloats}
\usepackage{multirow}
\usepackage{url}

\newcommand{\TODO}[1]{\textbf{\color{red}[TODO: #1]}}

\renewcommand{\TODO}[1]{}

\definecolor{cvprblue}{rgb}{0.21,0.49,0.74}
\usepackage[pagebackref,breaklinks,colorlinks,allcolors=cvprblue]{hyperref}

\title{An Empirical Study of Monocular Human Body Measurement Under Weak Calibration}

\author{
Gaurav Sekar\\
Independent Researcher\\
{\tt\small gaurav3594@gmail.com} \\[2ex]
{\small \url{https://github.com/Infonioknight/Monocular-Body-Measurement}}
}

\begin{document}
\maketitle
\begin{abstract}
Estimating human body measurements from monocular RGB imagery remains challenging due to scale ambiguity, viewpoint sensitivity, and the absence of explicit depth information. This work presents a systematic empirical study of three weakly calibrated monocular strategies: landmark-based geometry, pose-driven regression, and object-calibrated silhouettes, evaluated under semi-constrained conditions using consumer-grade cameras. Rather than pursuing state-of-the-art accuracy, the study analyzes how differing calibration assumptions influence measurement behavior, robustness, and failure modes across varied body types. The results reveal a clear trade-off between user effort during calibration and the stability of resulting circumferential quantities. This paper serves as an empirical design reference for lightweight monocular human measurement systems intended for deployment on consumer devices.
\end{abstract}
\section{Introduction}
\label{sec:intro}

The problem addressed in this work is the estimation of major human body measurements, including shoulder width, waist circumference, torso height, and leg length, from monocular RGB input captured using consumer-grade cameras. Such measurements are of interest in applications ranging from apparel sizing to fitness and human–computer interaction. We focus on settings where a single stationary camera is placed directly in front of the subject and captures the full body from head to toe. While monocular setups are attractive due to their accessibility and minimal hardware requirements, they inherently suffer from scale ambiguity and the absence of explicit depth information, making reliable measurement estimation challenging across different camera qualities and resolutions.

Rather than assuming specialized hardware, calibrated camera parameters, or depth sensors, this work studies approaches designed to operate under weak calibration assumptions typical of everyday devices. The considered setting is semi-constrained: the camera is approximately frontal and stationary, and the subject is fully visible in the frame. While the proposed approaches are designed to tolerate variations in camera resolution and minor viewpoint deviations, significant camera skew, off-axis perspectives, or highly unconstrained capture conditions are outside the scope of this work due to the fundamental limitations of monocular vision.

Monocular body measurement is inherently challenging because three-dimensional body proportions must be inferred from a two-dimensional image projection. The lack of explicit depth cues introduces scale ambiguity and limits the reliability of direct geometric reasoning. These challenges are further compounded by the need to operate across consumer devices with varying camera resolutions and imaging characteristics, making it difficult to design a standardized and reliable measurement pipeline without access to calibrated hardware. In practice, additional sources of error arise from viewpoint sensitivity, improper subject positioning, partial body cropping, and clothing effects—particularly loose-fitting garments that obscure true body contours and alter silhouette-based cues.

In this work, we conduct a systematic empirical study of weakly calibrated design choices for monocular body measurement. We examine how different calibration strategies and design decisions influence measurement stability and sensitivity to viewpoint variation when operating under the described capture conditions.

Through controlled evaluation on video captures, we compare representative approaches that differ in their calibration source and geometric assumptions, focusing on relative error behavior, robustness to subject motion, and common failure modes. This study provides a unified empirical reference for estimating human body proportions from monocular RGB imagery under semi-constrained capture conditions.
\section{Literature Review}
\subsection{Image-Based Anthropometric Estimation}

Early work on monocular anthropometric estimation primarily relied on explicit geometric reasoning and regression-based mappings from image-derived measurements to physical body dimensions. Single-camera, multi-view approaches using linear regression demonstrated that body measurements such as height and limb dimensions can be recovered when multiple viewpoints or constrained capture protocols are available \cite{liu2018single}. Related studies extended these ideas to predicting body mass index (BMI) and regional body part sizes from multi-view body images using statistical learning models \cite{kim2023multi}.

While these methods establish the feasibility of image-based anthropometry, they typically assume strictly controlled camera placement, strictly stable subject pose, or multiple views, limiting their applicability under weakly calibrated, consumer-grade monocular capture conditions.

\subsection{Learning-Based Pose and Shape Models}

Recent advances in computer vision have emphasized learning-based recovery of three-dimensional human pose and body shape from monocular RGB images. Parametric body models such as SMPL \cite{loper2015smpl}, popularized by work from Kanazawa et al. \cite{pavlakos2018estimate}, leverage strong statistical priors learned from large-scale datasets to infer full 3D meshes. These approaches achieve compelling pose and shape reconstructions and have been widely adopted in graphics and vision applications.

However, such models are optimized for pose and shape estimation rather than direct anthropometric measurement. Metric scale is often implicit or normalized, and reliable physical measurements typically require additional calibration or assumptions. As a result, these methods address a different problem setting than lightweight measurement pipelines designed for weakly calibrated consumer devices.

\subsection{Consumer-Grade and Webcam-Based Measurement Systems}

Several studies have focused on practical body measurement systems using webcams or mobile devices. Early webcam-based approaches demonstrated the accessibility of monocular measurement pipelines but relied on constrained capture conditions and heuristic scaling rules \cite{gan2020anthrowebcam}. Other application-oriented systems, such as FITME and MediaPipe-based measurement pipelines, integrate pose estimation with machine learning or rule-based geometry to produce end-to-end anthropometric measurements for fashion and fitness use cases \cite{ashmawi2019fitme, pabba2025automatic}.

These systems highlight the practical relevance of monocular anthropometry on consumer hardware. However, they are typically evaluated as complete applications and do not isolate the impact of individual design choices, such as calibration strategy, landmark stability, or viewpoint sensitivity, on measurement accuracy.

\subsection{Pose Landmarks for Anthropometric Measurement}

A number of approaches explicitly utilize pose landmarks as proxies for anthropometric measurements. MediaPipe-based pipelines and convolutional landmark detectors extract linear distances between detected keypoints and apply geometric or regression-based transformations to estimate physical body dimensions \cite{zong2023mediapipe, wang2020resnet}. These methods demonstrate that sparse 2D landmarks can support approximate measurement without explicit 3D reconstruction.

At the same time, landmark accuracy is sensitive to pose variation, viewpoint, and occlusion. Clinical and kinematic validation studies evaluating monocular pose estimation against reference systems show that even when average accuracy is high, joint localization errors persist and vary across body parts and motion conditions \cite{Rode2025AssessmentOM, stam2020accuracy}. Landmark errors propagate to scale-calibrated measurements, motivating the use of temporal consistency constraints and conservative confidence thresholds in measurement pipelines.

\subsection{Silhouette-Based Anthropometry and Weak Calibration}

An alternative line of research exploits body silhouettes and contour geometry rather than sparse landmarks. Xie et al. demonstrated that whole-body silhouettes alone encode rich anthropometric and body composition information, enabling accurate estimation of fat and lean mass using contour-based shape models validated against DXA ground truth \cite{xie2015dxa}. These results highlight the representational power of silhouette geometry without requiring explicit three-dimensional reconstruction.

From a geometric perspective, the inherent scale ambiguity of monocular vision has long been studied in single-view metrology. Criminisi et al. showed that metric measurements from a single image are only possible through additional constraints, such as known reference dimensions or planar assumptions \cite{criminisi2000single}. While silhouette-based approaches demonstrate strong potential, most prior work operates under controlled acquisition pipelines or specialized imaging conditions, leaving open questions regarding robustness under weak calibration and consumer-grade capture.

\subsection{Positioning of the Present Work}

In contrast to prior work that assumes strong calibration, focuses on end-to-end application systems, or prioritizes full 3D reconstruction, this study presents a systematic empirical analysis of weakly calibrated monocular body measurement strategies. By comparing landmark-based calibration, pose-driven regression, and object-calibrated, silhouette-assisted estimation within a unified evaluation pipeline, this work isolates how calibration strength, geometric cues, and temporal consistency influence measurement stability under semi-constrained, consumer-grade capture conditions.
\section{Evaluation Setup}
\subsection{Capture Setup}

All video captures were obtained using the built-in webcam of a consumer-grade laptop (MacBook Air M1). Recordings were performed indoors under typical ambient lighting conditions. The camera was placed on a stationary surface at a fixed height and oriented to capture a frontal view of the subject, with the full body visible from head to toe. Camera resolution was not fixed and varied according to default device settings.

\subsection{Subjects and Ground Truth}

The evaluation was conducted on four adult subjects, including the author and household members. The subjects span a wide range of ages (23–78 years) and exhibit varied body types. No explicit selection criteria were applied beyond adult age, as the goal of the evaluation was to assess behavior across a small but diverse set of physiques rather than to establish population-level generalization.

Ground-truth body measurements were obtained manually using a standard tape measure. Measurements were taken following a consistent procedure across all subjects and recorded in inches. For each subject, ground-truth values were collected once for the relevant body dimensions, and these measurements are treated as reference values for computing absolute error. All subjects were measured under the same protocol to ensure consistency.

\subsection{Measurement Protocol}

For each subject, body measurements were obtained from a single video capture per method. All methods operate on a frontal monocular video sequence in which the subject stands upright and fully visible within the frame. The same frontal input sequence is used for both the anthropometric calibration baseline and the pose-based regression approach to ensure a consistent basis for comparison. The object-calibrated silhouette-based method additionally requires a separate side-view capture, which is used exclusively for estimating lateral body extent.

Calibration is performed once per capture session. In the first two methods, a single calibration step is used to recover a global pixel-to-metric scale, which is then reused for all subsequent measurements within the session under the assumption of fixed camera placement and subject positioning. In the final method, metric scale is recovered through object-based calibration and refined using a human-in-the-loop height correction, without requiring explicit reuse of calibration state across captures.

Across all methods, a temporal consistency constraint is applied to ensure reliable measurement extraction. A frame is considered valid only if all required pose landmarks are detected with sufficient confidence, and measurements are extracted only after this condition is satisfied for a fixed number of consecutive frames. Once the temporal criterion is met, body measurements are computed from the final valid frame in the sequence.

\subsection{Metrics}

Evaluation is performed using absolute measurement error between estimated values and tape-measured ground truth. All measurements are reported in inches. We evaluate shoulder width, waist circumference, torso height, and leg length, as these quantities are either directly estimated or can be meaningfully derived across all considered methods.

Body height is excluded from evaluation, as it is used as an input during calibration in the object-calibrated method and therefore does not constitute an independent prediction. Circumferential measurements predicted exclusively by the regression-based approach but not estimated by other methods are similarly excluded to ensure fair comparison. 

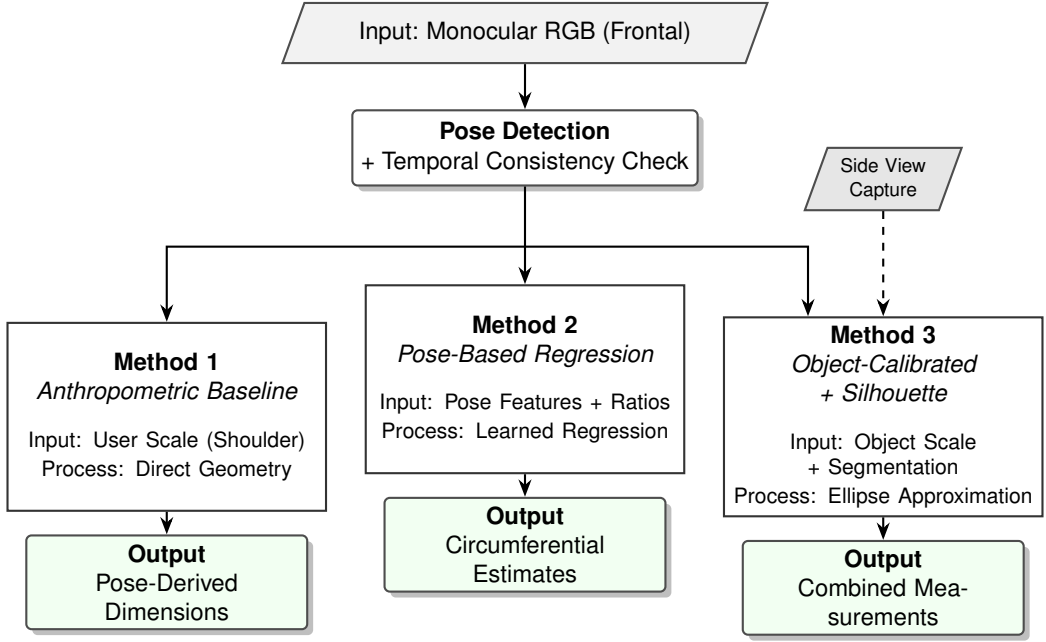
\begin{figure*}[t]
\centering
\begin{tikzpicture}[
    node distance=1.5cm and 0.5cm,
    font=\sffamily\small,
    >=Stealth,
    block/.style={
        rectangle, 
        draw=black!70, 
        fill=white, 
        thick, 
        rounded corners=2pt, 
        align=center, 
        minimum height=1cm, 
        minimum width=3.5cm,
        drop shadow
    },
    input/.style={
        trapezium, 
        trapezium left angle=70, 
        trapezium right angle=110, 
        draw=black!70, 
        fill=gray!10, 
        thick, 
        align=center, 
        minimum height=0.8cm, 
        minimum width=2.5cm,
        inner sep=5pt
    },
    method/.style={
        rectangle, 
        draw=black!80, 
        fill=white, 
        thick, 
        align=center, 
        minimum height=2.5cm, 
        text width=4cm, 
        yshift=-0.5cm
    },
    line/.style={
        draw, 
        thick, 
        ->
    },
    dashedline/.style={
        draw, 
        thick, 
        dashed,
        ->
    }
]

    \node (video) [input] {Input: Monocular RGB (Frontal)};
    
    \node (proc) [block, below=0.6cm of video] {
        \textbf{Pose Detection} \\ 
        + Temporal Consistency Check
    };

    \coordinate [below=0.8cm of proc] (branch);

    
    \node (m2) [method, below=0.8cm of proc] {
        \textbf{Method 2}\\
        \textit{Pose-Based Regression}\\[0.3cm]
        \footnotesize Input: Pose Features + Ratios\\
        Process: Learned Regression
    };

    \node (m1) [method, left=0.5cm of m2] {
        \textbf{Method 1}\\
        \textit{Anthropometric Baseline}\\[0.3cm]
        \footnotesize Input: User Scale (Shoulder)\\
        Process: Direct Geometry
    };

    \node (m3) [method, right=0.5cm of m2] {
        \textbf{Method 3}\\
        \textit{Object-Calibrated + Silhouette}\\[0.3cm]
        \footnotesize Input: Object Scale + Segmentation\\
        Process: Ellipse Approximation
    };

    \node (sideview) [input, above=1.4cm of m3, scale=0.8, text width=1.5cm, fill=gray!20] {Side View Capture};

    \node (out1) [block, below=0.3cm of m1, fill=green!5, minimum height=0.8cm, text width=3.5cm] {
        \textbf{Output}\\
        Pose-Derived Dimensions
    };
    
    \node (out2) [block, below=0.3cm of m2, fill=green!5, minimum height=0.8cm, text width=3.5cm] {
        \textbf{Output}\\
        Circumferential Estimates
    };

    \node (out3) [block, below=0.3cm of m3, fill=green!5, minimum height=0.8cm, text width=3.5cm] {
        \textbf{Output}\\
        Combined Measurements
    };

    \draw [line] (video) -- (proc);
    \draw [draw, thick] (proc) -- (branch);
    
    \draw [line] (branch) -| (m1.north);
    \draw [line] (branch) -- (m2.north);
    
    \draw [line] (branch) -| ([xshift=-1.0cm]m3.north);

    \draw [dashedline] (sideview.south) -- (m3.north);

    \draw [line] (m1.south) -- (out1.north);
    \draw [line] (m2.south) -- (out2.north);
    \draw [line] (m3.south) -- (out3.north);

\end{tikzpicture}
\caption{Overview of the evaluation pipeline}
\label{fig:method_pipeline}
\end{figure*}

Absolute error is reported per subject and per method, without aggregation across subjects.

\section{Methodology}

All evaluated approaches operate on monocular RGB video captured under a semi-constrained frontal setup. Pose landmarks are extracted continuously from each frame, and a frame is considered valid only if all required landmarks are detected with a visibility confidence exceeding 0.9. To improve robustness to transient detection errors and motion blur, measurements are extracted only after these conditions are satisfied for a fixed number of consecutive frames, after which measurements are computed from the final frame in the sequence. These preprocessing and filtering steps are shared across all evaluated methods unless otherwise stated.

\subsection{Overview}

We evaluate three representative approaches for monocular body measurement that differ primarily in their calibration source and degree of geometric reasoning. An overview of the evaluation pipeline is shown in Fig.~\ref{fig:method_pipeline}.

The first approach serves as a minimal anthropometric baseline, in which metric scale is established through a single user-provided body reference and subsequently applied to pose-derived measurements.

The second approach augments this baseline with learned pose-driven priors to estimate circumferential quantities from two-dimensional landmark information, introducing stronger statistical assumptions in place of purely geometric scaling.

The third approach adopts an object-calibrated strategy that incorporates silhouette-based cues and an auxiliary side-view capture to indirectly reason about body depth, enabling more stable estimation of three-dimensional measurements under favorable capture conditions.

\subsection{Anthropometric Calibration Baseline}

The first approach serves as a simple anthropometric calibration baseline. Calibration is performed once during an initial setup step using a user-provided shoulder-width measurement. Specifically, the pixel distance between the left and right shoulder landmarks is measured. This value is then compared against the user-provided shoulder width in inches to derive a global pixel-to-metric scaling factor. This scaling factor is then reused for all subsequent measurements without re-calibration, and all remaining body dimensions are computed directly from pose-derived pixel distances using this fixed scale.

This baseline implicitly assumes consistent camera placement and subject positioning across captures. Subjects were instructed to maintain consistent positioning during calibration, reflecting the assumption that body pose and camera geometry remain stable once calibration is completed. As a result, this approach is highly sensitive to subject movement, viewpoint changes, and inter-subject anatomical variability. Despite its limitations, this method provides a minimal reference point against which the robustness of more advanced calibration strategies can be evaluated.

\subsection{Pose-Based Regression for Circumferential Estimation}

The second approach extends the anthropometric calibration baseline by introducing learned regression models for estimating circumferential body measurements. In this approach, a set of pose-derived anthropometric measurements is extracted. In addition to these measurements, a simple ratio feature representing the torso-to-leg length is computed from pose landmarks. These quantities are used as inputs to two separate regression models trained to predict waist and chest circumference, respectively. The regression models are trained on an external body proportion dataset (BodyM) \cite{ruiz2022humanbodymeasurementestimation}, enabling indirect estimation of circumferential measurements that are difficult to obtain reliably from frontal pose geometry alone.

While this approach allows circumferential quantities to be inferred without explicit depth information, it remains dependent on the accuracy and stability of pose landmarks and the assumptions implicit in the training data. As with the baseline, the method does not explicitly model three-dimensional shape or depth and remains sensitive to pose variation and viewpoint changes.

\subsection{Object-Calibrated Silhouette-Assisted Estimation}

The final approach replaces the purely pose-based calibration strategy with an object-calibrated, silhouette-assisted measurement pipeline designed to improve robustness to scale ambiguity and circumferential estimation.

Metric scale is recovered using a known-size reference object placed in the scene. The person and reference object are detected and segmented, and the contour of the reference object is used to estimate a pixel-to-metric scaling factor by comparing its measured extent against its known physical dimensions. This scale is applied to all subsequent measurements.

Using the recovered scale, an initial estimate of the subject’s height is obtained from the segmented person mask in the frontal view. To mitigate residual errors due to vertical perspective distortion, this estimate is optionally corrected by the user in a human-in-the-loop step. This correction refines the height estimate only and does not overwrite or replace the object-based scale recovery.

Body measurements are computed by combining pose landmarks and segmentation cues. Linear dimensions such as shoulder width, torso height, and leg length are derived from the frontal pose landmarks. To better estimate circumferential quantities, an additional side-view capture is incorporated, from which the segmented person mask is used to estimate the extent of the lateral body.

Circumferential measurements such as waist are approximated by combining frontal and side-view extents under a simple geometric assumption, modeling the torso cross-section as an ellipse. Given a frontal width $w_f$ and a lateral width $w_s$, waist circumference $C$ is estimated as
\[
C \approx \pi \left[ 3(a + b) - \sqrt{(3a + b)(a + 3b)} \right],
\]
where $a = \tfrac{w_f}{2}$ and $b = \tfrac{w_s}{2}$. While this approximation does not capture true three-dimensional body shape, it provides a lightweight means of incorporating depth-related information using only monocular inputs.
\section{Results}

\subsection{Per-Method Measurement Accuracy}

Table \ref{tab:empirical_results} reports estimated body measurements for each subject alongside tape-measured ground truth. Results are presented per subject without aggregation to emphasize method-level behavior under identical capture conditions. Shoulder width is included for completeness but is not interpreted as a predictive result for Methods 1 and 2, as it is used directly for scale calibration.

\begin{table*}[t]
\centering
\caption{Empirical body measurements across evaluated methods (inches).}
\label{tab:empirical_results}
\small
\begin{tabular}{l p{1.6cm} c c c c}
\toprule
\textbf{Subject} & \textbf{Method} & \textbf{Shoulder*} & \textbf{Torso} & \textbf{Waist} & \textbf{Leg Height} \\
\midrule
\multirow{4}{*}{S1}
 & GT & 15.00 & 20.00 & 29.00 & 41.00 \\
 & M1 & 15.00 & 21.22 & 31.55 & 35.57 \\
 & M2 & 15.00 & 21.22 & 29.13 & 35.57 \\
 & M3 & 14.48 & 21.04 & 37.08 & 39.17 \\
\midrule
\multirow{4}{*}{S2}
 & GT & 15.00 & 18.00 & 37.50 & 41.00 \\
 & M1 & 15.00 & 20.87 & 30.41 & 30.76 \\
 & M2 & 15.00 & 20.87 & 27.73 & 30.76 \\
 & M3 & 14.52 & 20.48 & 37.59 & 35.65 \\
\midrule
\multirow{4}{*}{S3}
 & GT & 12.00 & 15.00 & 29.50 & 31.00 \\
 & M1 & 12.00 & 16.75 & 25.55 & 27.92 \\
 & M2 & 12.00 & 16.75 & 21.06 & 27.92 \\
 & M3 & 11.58 & 16.13 & 31.56 & 30.20 \\
\midrule
\multirow{4}{*}{S4}
 & GT & 13.50 & 16.50 & 33.00 & 34.50 \\
 & M1 & 13.50 & 18.25 & 25.83 & 30.14 \\
 & M2 & 13.50 & 18.25 & 29.97 & 30.14 \\
 & M3 & 13.32 & 17.64 & 34.16 & 32.67 \\
\bottomrule
\multicolumn{6}{l}{\footnotesize *M1 and M2 use ground-truth shoulder width for scale calibration.}
\end{tabular}
\end{table*}

The anthropometric baseline (M1) produces reasonable estimates for linear dimensions such as torso and leg height but exhibits substantial variability for circumferential quantities. This behavior is expected given the reliance on a single global scaling factor and the absence of depth-related cues.

The pose-based regression approach (M2) does not consistently outperform the baseline across subjects. While modest improvements are observed for waist estimation in some cases, performance remains variable and sensitive to pose-derived feature stability. In several instances, M2 produces circumferential estimates comparable to or worse than M1, indicating that regression-based correction alone is insufficient to reliably resolve depth ambiguity under the evaluated weak calibration conditions.

The object-calibrated silhouette-assisted method (M3) yields the most accurate circumferential estimates when the silhouette assumptions are satisfied. By incorporating both frontal and lateral extent through an explicit geometric approximation, M3 produces waist measurements that are substantially closer to ground truth for subjects whose silhouette assumptions are satisfied. Linear measurements such as leg height also show improved consistency relative to pose-only methods.

\subsection{Comparison Across Calibration Strategies}

Absolute measurement error for all methods is summarized in Table \ref{tab:absolute_error}. Errors are reported in inches to preserve physical interpretability and to avoid misleading normalization effects for small sample sizes.

Increasing calibration strength does not lead to uniform improvement across all measurements. Instead, gains are measurement-dependent. Object-based calibration provides clear benefits for leg height and circumferential estimation, while torso height shows more modest improvements. Pose-only methods remain sensitive to scale drift and landmark variability, particularly for measurements that implicitly depend on body depth.

Circumferential estimation benefits most from the object-calibrated strategy. Unlike M1 and M2, which implicitly assume a fixed proportional relationship between frontal width and true circumference, M3 explicitly models lateral body extent and applies a more anatomically plausible approximation. This results in substantially lower waist error for subjects whose clothing and pose preserve a faithful silhouette.

These results indicate that stronger calibration trades increased setup complexity for improved robustness, particularly when estimating measurements that cannot be reliably inferred from frontal pose geometry alone.

\begin{table*}[t]
\centering
\caption{Absolute measurement error with respect to ground truth (inches).}
\label{tab:absolute_error}
\small
\begin{tabular}{l p{1.6cm} c c c c}
\toprule
\textbf{Subject} & \textbf{Method} & \textbf{Shoulder} & \textbf{Torso} & \textbf{Waist} & \textbf{Leg Height} \\
\midrule
\multirow{3}{*}{S1}
 & M1 & 0.00 & 1.22 & 2.55 & 5.43 \\
 & M2 & 0.00 & 1.22 & \textbf{0.13} & 5.43 \\
 & M3 & 0.52 & \textbf{1.04} & 8.08 & \textbf{1.83} \\
\midrule
\multirow{3}{*}{S2}
 & M1 & 0.00 & 2.87 & 7.09 & 10.24 \\
 & M2 & 0.00 & 2.87 & 9.77 & 10.24 \\
 & M3 & 0.48 & \textbf{2.48} & \textbf{0.09} & \textbf{5.35} \\
\midrule
\multirow{3}{*}{S3}
 & M1 & 0.00 & 1.75 & 3.95 & 3.08 \\
 & M2 & 0.00 & 1.75 & 8.44 & 3.08 \\
 & M3 & 0.42 & \textbf{1.13} & \textbf{2.06} & \textbf{0.80} \\
\midrule
\multirow{3}{*}{S4}
 & M1 & 0.00 & 1.75 & 7.17 & 4.36 \\
 & M2 & 0.00 & 1.75 & 3.03 & 4.36 \\
 & M3 & 0.18 & \textbf{1.14} & \textbf{1.16} & \textbf{1.83} \\
\bottomrule
\multicolumn{6}{l}{\footnotesize *Zero shoulder error for M1 and M2 reflects calibration, not predictive accuracy.}
\end{tabular}
\end{table*}

\subsection{Failure Modes and Qualitative Observations}

\begin{figure}[tb]
  \centering
  \includegraphics[width=\columnwidth]{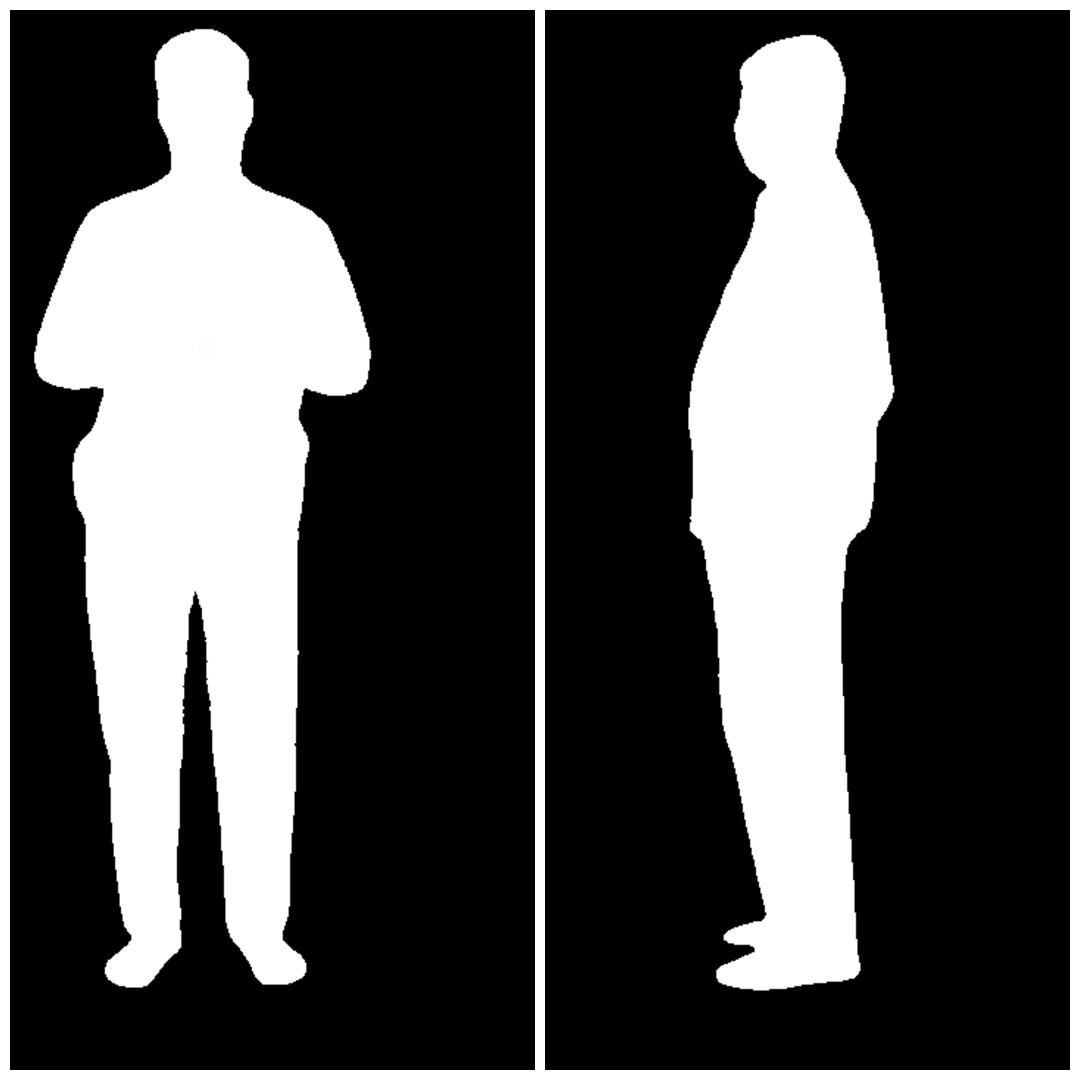}
  \caption{Frontal and lateral person silhouettes extracted under fitted clothing conditions, where clothing preserves body contour.}
  \label{fig:silhouette_fitted}
\end{figure}

The largest circumferential errors observed for the object-calibrated method occur when silhouette assumptions are violated. In particular, loose-fitting clothing inflates apparent body width in segmentation masks, leading to systematic overestimation of waist circumference. This effect is most pronounced for Subject 1, who wore a loose garment during capture. Figure~\ref{fig:silhouette_fitted} shows a fitted-clothing case in which the extracted frontal and lateral silhouettes closely follow true body contour.

Conversely, subjects wearing well-fitted clothing exhibit substantially more accurate circumferential estimates, even in the absence of tight or compression garments. Minor deviations are also observed when clothing fit varies across body regions, such as increased looseness around the waist due to upper-body prominence.

These errors arise when the silhouette deviates from true body contour rather than from instability in the measurement pipeline itself. Such limitations are inherent to silhouette-based anthropometric methods and reflect a fundamental trade-off between accessibility and measurement fidelity under weak calibration and monocular capture.
\section{Discussion and Limitations}

This work presents an empirical comparison of three monocular human body measurement strategies that vary in complexity, calibration strength, and underlying assumptions. Rather than proposing a new model, the goal is to understand how different design choices behave in practice when estimating linear and circumferential body measurements from monocular RGB input under weak calibration assumptions.

The evaluation is intentionally conducted on a small number of subjects. While the sample size is limited, the subjects differ substantially in body type, age, and appearance, which allows the study to expose the conditions under which each method performs well and where it fails. Despite the small N, the observed trends highlight relative strengths, weaknesses, and feasibility trade-offs across the evaluated approaches. The study is therefore best interpreted as a behavior-focused evaluation rather than a population-level validation.

All methods assume a semi-constrained capture setup. Multi-subject scenes are explicitly unsupported and cause failure across the pipeline. Adequate lighting is also required, particularly for the object-calibrated method, as reference object detection and segmentation degrade significantly in low-light conditions. In addition, the reference object must be held in a consistent orientation such that its edges are clearly visible, and sufficient color contrast between the object, clothing, and background is necessary for reliable detection.

Clothing is the most influential factor affecting the performance of the object-calibrated silhouette-based method. Loose-fitting garments produce silhouettes that substantially overestimate true body width, leading to inflated circumferential measurements. This effect is clearly observed in Subject 1, whereas Subjects 2–4, who wore reasonably fitted clothing, exhibit much more accurate estimates. This sensitivity is inherent to silhouette-based reasoning rather than an implementation flaw. When silhouette assumptions are satisfied, the method provides a reasonable approximation of three-dimensional body extent despite operating entirely under monocular capture.

The evaluated methods differ not only in their geometric reasoning but also in their calibration source. Methods 1 and 2 rely on a user-provided body measurement for scale normalization, while Method 3 uses a rigid reference object present in the scene. This distinction is intentional and reflects a user-experience trade-off rather than an uncontrolled experimental variable. The comparison therefore evaluates two classes of weak calibration workflows: human-seeded calibration versus environment-seeded calibration. The observed differences in stability should be interpreted in this context.

Finally, each method exhibits a distinct trade-off profile. The anthropometric baseline operates purely in two dimensions and can produce reasonable linear estimates, but is extremely sensitive to user positioning due to its calibration strategy. The pose-based regression approach remains limited by the quality and stability of pose landmarks and does not reliably resolve depth ambiguity. The object-calibrated method introduces additional user effort and requires appropriate lighting, contrast, and clothing choices. These limitations reflect practical trade-offs rather than implementation flaws and should be considered when selecting or deploying lightweight monocular body measurement systems.

\section{Conclusion}

This paper presented a systematic empirical evaluation of three monocular human body measurement approaches operating under weak calibration assumptions. By comparing a landmark-based anthropometric baseline, a pose-driven regression strategy, and an object-calibrated silhouette-assisted method within a unified pipeline, the study examined how increasing calibration strength and geometric reasoning affects the estimation of both linear and circumferential body measurements.

The results show that purely pose-based methods are capable of producing reasonable two-dimensional measurements but are highly sensitive to calibration and positioning, particularly when estimating quantities that implicitly depend on body depth. Incorporating silhouette cues and object-based scale information enables more reliable approximation of three-dimensional body extent, provided that key assumptions regarding clothing, lighting, and visibility are satisfied. At the same time, the study highlights clear failure modes and practical constraints that arise when operating under monocular capture without calibrated hardware or depth sensors. 

Overall, this work serves as an empirical reference that consolidates and contrasts several relevant monocular measurement strategies, clarifies their limitations, and highlights the trade-offs involved in designing accessible, lightweight human body measurement systems. Future work could explore strategies that reduce silhouette sensitivity to clothing while preserving the accessibility and low calibration overhead of monocular capture.

{
    \small
    \bibliographystyle{ieeenat_fullname}
    \bibliography{main}
}
\end{document}